\documentclass[11pt,a4paper]{article}
\usepackage[hyperref]{emnlp2020}
\usepackage{times}
\usepackage{latexsym}

\usepackage{microtype}

\usepackage{url}

\usepackage{graphicx}
\usepackage{CJKutf8}
\usepackage{multirow}
\usepackage{array}
\usepackage{tabularx}
\usepackage{makecell}
\usepackage{amsmath} 
\usepackage{amssymb}
\usepackage{tablefootnote}
\usepackage{subcaption}
\usepackage{diagbox}
\usepackage{algorithm}
\usepackage[noend]{algpseudocode}
\usepackage{booktabs}
\usepackage{xcolor}
\usepackage{color, soul} 
\usepackage{ccg-latex}

\usepackage{microtype}

\newcolumntype{L}[1]{>{\raggedright\arraybackslash}p{#1}}
\newcolumntype{C}[1]{>{\centering\arraybackslash}p{#1}}
\newcolumntype{R}[1]{>{\raggedleft\arraybackslash}p{#1}}

\DeclareSymbolFont{extraup}{U}{zavm}{m}{n}
\DeclareMathSymbol{\varheart}{\mathalpha}{extraup}{86}
\DeclareMathSymbol{\vardiamond}{\mathalpha}{extraup}{87}

\aclfinalcopy 
\def\aclpaperid{2529} 

\title{Supertagging Combinatory Categorial Grammar\\with Attentive Graph Convolutional Networks}

\author{
    Yuanhe Tian$^{\varheart}$, \hspace{0.2cm}
    Yan Song$^{{\spadesuit}\heartsuit\dag}$, \hspace{0.2cm}
    Fei Xia$^{\varheart}$\\
    $^{\varheart}$University of Washington \hspace{0.4cm}
    $^{\spadesuit}$The Chinese University of Hong Kong (Shenzhen)\\
    $^{\heartsuit}$Shenzhen Research Institute of Big Data \\
    $^{\varheart}$\texttt{\{yhtian, fxia\}@uw.edu} \hspace{0.4cm}
    $^{\spadesuit}$\texttt{songyan@cuhk.edu.cn} \\ \hspace{0.4cm}
}

\date{}

\begin{document}
\maketitle

\renewcommand{\thefootnote}{\fnsymbol{footnote}}
\footnotetext[2]{Corresponding author.}

\renewcommand{\thefootnote}{\arabic{footnote}}

\begin{abstract}

Supertagging is conventionally regarded as an important task for combinatory categorial grammar (CCG) parsing, where effective modeling of contextual information is highly important to this task.
However, existing studies have made limited efforts to leverage contextual features except for applying powerful encoders (e.g., bi-LSTM).
In this paper, we propose attentive graph convolutional networks to enhance neural CCG supertagging through a novel solution of leveraging contextual information.
Specifically, we build the graph from chunks (n-grams) extracted from a lexicon and apply attention over the graph, so that different 
word pairs
from the contexts within and across chunks are weighted in the model and facilitate the supertagging accordingly.
The experiments performed on the CCGbank demonstrate that our approach outperforms all previous studies
in terms of both supertagging and parsing.
Further analyses illustrate the effectiveness of each component in our approach to discriminatively learn from word pairs to enhance CCG supertagging.\footnote{Our code and models for CCG supertagging are released at \url{https://github.com/cuhksz-nlp/NeST-CCG}.}

\end{abstract}

\section{Introduction} \label{intro}

%


Combinatory categorial grammar (CCG) is a lexicalized grammatical formalism, where the lexical categories (also known as supertags) of the words in a sentence provide informative syntactic and semantic knowledge for text understanding.
Therefore, CCG parse often provides useful information for many downstream natural language processing (NLP) tasks such as logical reasoning \cite{yoshikawa-etal-2018-consistent} and semantic parsing \cite{beschke-2019-exploring}.
To perform CCG parsing in different languages,
most studies conducted a supertagging-parsing pipline \cite{clark-curran-2007-wide,kummerfeld-etal-2010-faster,song2012construction,lewis-steedman-2014-improved,huang2015chinese, xu-etal-2015-ccg, lewis-etal-2016-lstm, vaswani-etal-2016-supertagging, yoshikawa-etal-2017-ccg}, in which their main focus is the first step, and they generated the CCG parse trees directly from supertags with a few rules afterwards.
%

Building an accurate supertagger in a sequence labeling process requires a good modeling of contextual information.
Recent neural approaches to supertagging mainly focused on leveraging powerful encoders with recurrent models \cite{lewis-etal-2016-lstm, vaswani-etal-2016-supertagging,clark2018semi}, with limited attention paid to modeling extra contextual features such as word pairs with strong relations.
Graph convolutional networks (GCN) is demonstrated to be an effective approach to model such contextual information between words in many NLP tasks \cite{marcheggiani-titov-2017-encoding, huang-carley-2019-syntax, de-cao-etal-2019-question,huang-etal-2019-text}; thus we want to determine whether this approach can 
also help CCG supertagging.

However, we cannot directly apply conventional GCN models to CCG supertagging
because in most of the previous studies the GCN models are built over the edges
in the dependency tree of an input sentence.
As high-quality dependency parsers are not always available, we do not want our CCG supertaggers to rely on the existence of dependency parsers. 
Thus, we need another way to extract useful word pairs to build GCN models.
For that, we propose to obtain word pairs from frequent chunks (n-grams) in the corpus, because those chunks are easy to identify with co-occurrence counts.
To appropriately learn from n-grams, one requires the GCN to be able to distinguish different word pairs because such information in n-grams are not explicitly structured as that in dependency parses.
Because existing GCN models are limited in treating all word pairs equally,
we propose an adaptation of conventional GCN for CCG supertagging.

\begin{figure}[t]
    \centering
    \includegraphics[width=0.48\textwidth, trim=0 15 0 0]{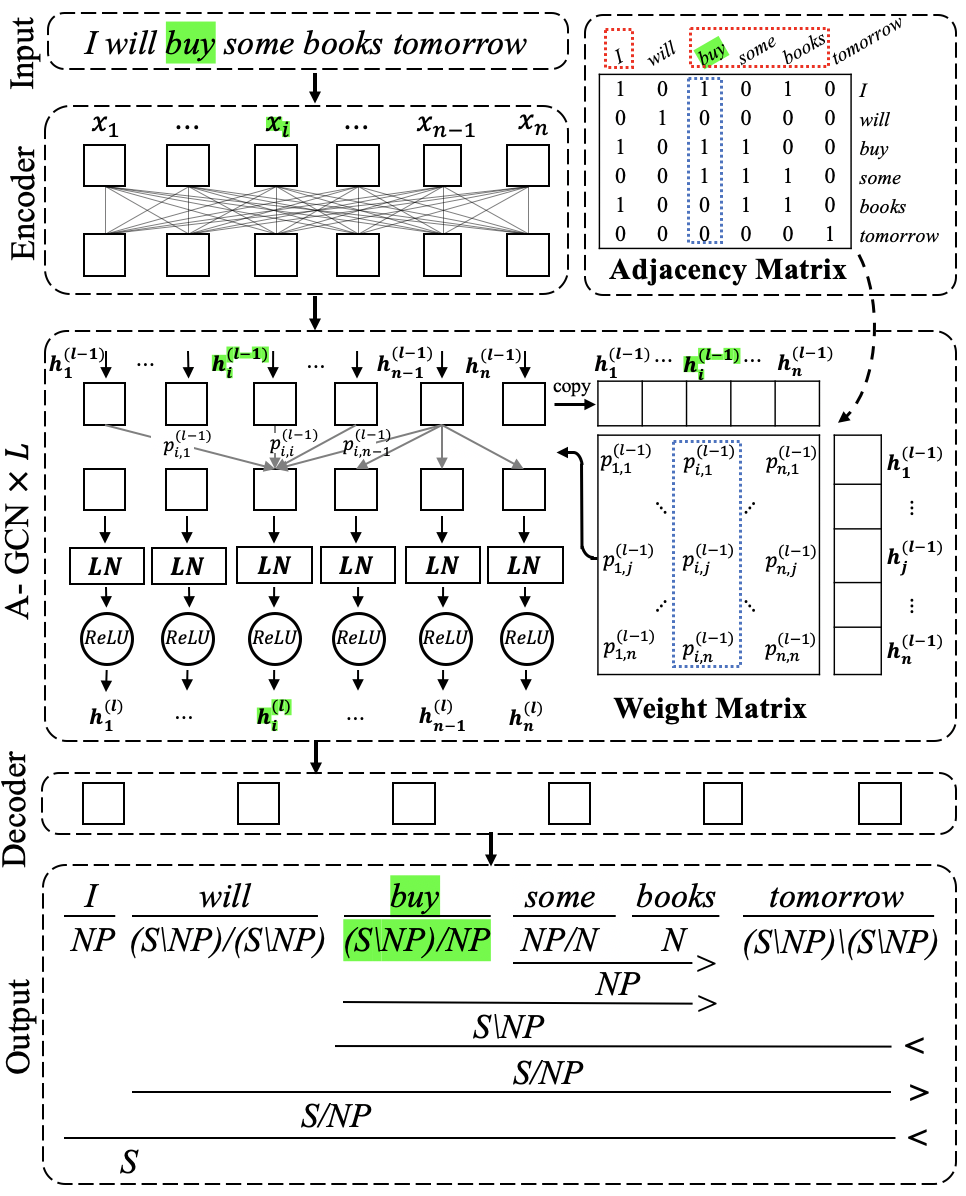}
    \caption{The architecture of our CCG supertagger with A-GCN and an example input sentence with its supertagging and parsing output. The supertagging process for ``\textit{buy}'' is highlighted in green.
    The adjacency matrix illustrates the edges of the graph that is built upon the chunks (n-grams) extracted from the lexicon $\mathcal{N}$, with the chunks illustrated in the red boxes.}
    \label{fig:model}
    \vskip -1em
\end{figure}

In this paper, we propose attentive GCN (A-GCN) for CCG supertagging, where its input graph is built based on chunks (n-grams) extracted with unsupervised methods.
In detail, two types of edges in the graph are introduced to model word relations within and across chunks
%
and an attention mechanism is applied to GCN to weight those edges.
%
In doing so, different contextual information are discriminatively learned to facilitate CCG supertagging without requiring any external resources.
%
The validity of our approach is demonstrated by experimental results on the CCGbank \cite{hockenmaier-steedman-2007-ccgbank}, where state-of-the-art performance is obtained for both tagging and parsing.

\section{The Approach}

We treat CCG supertagging as a sequence labeling task,
where the input is a sentence with $n$ words $\mathcal{X}=x_{1}x_{2} \cdots x_{i} \cdots x_{n}$, 
and the output is 
a sequence of supertags $\widehat{\mathcal{Y}}=\widehat{y}_{1}\widehat{y}_{2} \cdots \widehat{y}_{i} \cdots \widehat{y}_{n}$.
Our approach uses attentive GCN (A-GCN) to incorporate information of word pairs through a graph;
the graph is built based on n-grams in the input sentence that appear in a lexicon $\mathcal{N}$.
This lexicon consists of 
n-grams automatically extracted from raw corpora by unsupervised methods.
%
%
The overall architecture of our tagger is illustrated in Figure \ref{fig:model}, with an input sentence and corresponding supertagging and parsing output.
%
%
The details of the main components in the architecture are provided below.

\subsection{GCN}

%
Normal GCN models with $L$ layers learn from word pairs suggested by the dependency parsing results of the input sentence $\mathcal{X}$, where the edges between all pairs of words $x_i$ and $x_j$ are represented by an adjacency matrix $\mathcal{A}=\{a_{i,j}\}_{n \times n}$.
In $\mathcal{A}$, $a_{i,j}=1$ if there is a dependency edge between $x_i$ and $x_j$ or $i=j$
(the direction of the edge is ignored), and $a_{i,j}=0$ otherwise.
Based on the adjacency matrix, for each $x_i$, the $l$-th GCN layer finds all $x_j$ associated with $x_i$ (where $a_{i,j}=1$), takes their hidden vectors $\mathbf{h}^{(l-1)}_j$ from the $(l-1)$-th layer, and computes the output for $x_i$ by 
\begin{equation} \label{eq: gcn}
\setlength\abovedisplayskip{6pt}
\setlength\belowdisplayskip{6pt}
    \mathbf{h}^{(l)}_{i} = \sigma (LN(\sum_{j=1}^{n} a_{i,j}
    (\mathbf{W}^{(l)} \cdot \mathbf{h}^{(l-1)}_{j} + \mathbf{b}^{(l)})))
\end{equation}
where $\mathbf{W}^{(l)}$ and $\mathbf{b}^{(l)}$ are trainable matrix and bias for the $l$-th GCN layer, $LN$ refers to layer normalization and $\sigma$ the $ReLU$ activation function.
Therefore, in normal GCN, for each $x_i$, all the $x_j$ that connect to $x_i$ are treated 
exactly the same.

\subsection{Graph Construction based on Chunks}

Since CCG supertagging is also a parsing task, we do not want our approach to rely on the
existence of a dependency parser.
Without such a parser, we need an alternative for finding good word pairs to build the graph in A-GCN (which is equivalent to build the adjacency matrix $\mathcal{A}$).
%
%
%
%
%
%
Inspired by the studies that leverage chunks (n-grams) as effective features to carry contextual information and enhance model performance \cite{song-etal-2009-transliteration,song2012using,ishiwatari-etal-2017-chunk,yoon-etal-2018-learning,zhang-etal-2019-incorporating,tian-etal-2020-joint,tian2020improving,tian-etal-2020-constituency}, we propose to construct the graph based on the chunks (n-grams) extracted from a pre-constructed n-gram lexicon $\mathcal{N}$.
%
%
Specifically, the lexicon is constructed by computing the PMI of any two adjacent words $s', s''$ in the training set by
\begin{equation}
\setlength\abovedisplayskip{6pt}
\setlength\belowdisplayskip{6pt}
    PMI(s', s'') = log \frac{p(s's'')}{p(s')p(s'')}
\end{equation}
where $p$ is the probability of an n-gram (i.e., $s'$, $s''$ and $s's''$) in the training set;
then a high PMI score suggests that the two words co-occur a lot in the dataset and are more likely to form a n-gram.
For each pair of adjacent words $s_{i-1}$, $s_{i}$ in a sentence $\mathcal{S}=s_1s_2 \cdots s_{i-1}s_{i} \cdots s_{n}$, we compute the PMI score of the two words and use a threshold to determine whether a delimiter should be inserted between them.
As a result, the sentence $\mathcal{S}$ is segmented into pieces of n-grams and we extract all n-grams from all sentences to form the lexicon $\mathcal{N}$.\footnote{For example, a sentence can be segmented into $\mathcal{S}=s_1/s_2s_3s_4/s_5$ (``/'' refers to a delimiter) if the PMI of $s_1, s_2$ and $s_4, s_5$ are lower than the threshold and the PMI of $s_2, s_3$ and $s_3, s_4$ are greater than the threshold; we thus obtain three n-grams, i.e., $s_1$, $s_2s_3s_4$, and $s_5$ from this sentence.}

\begin{figure}[t]
    \centering
    \includegraphics[width=0.48\textwidth, trim=0 20 0 0]{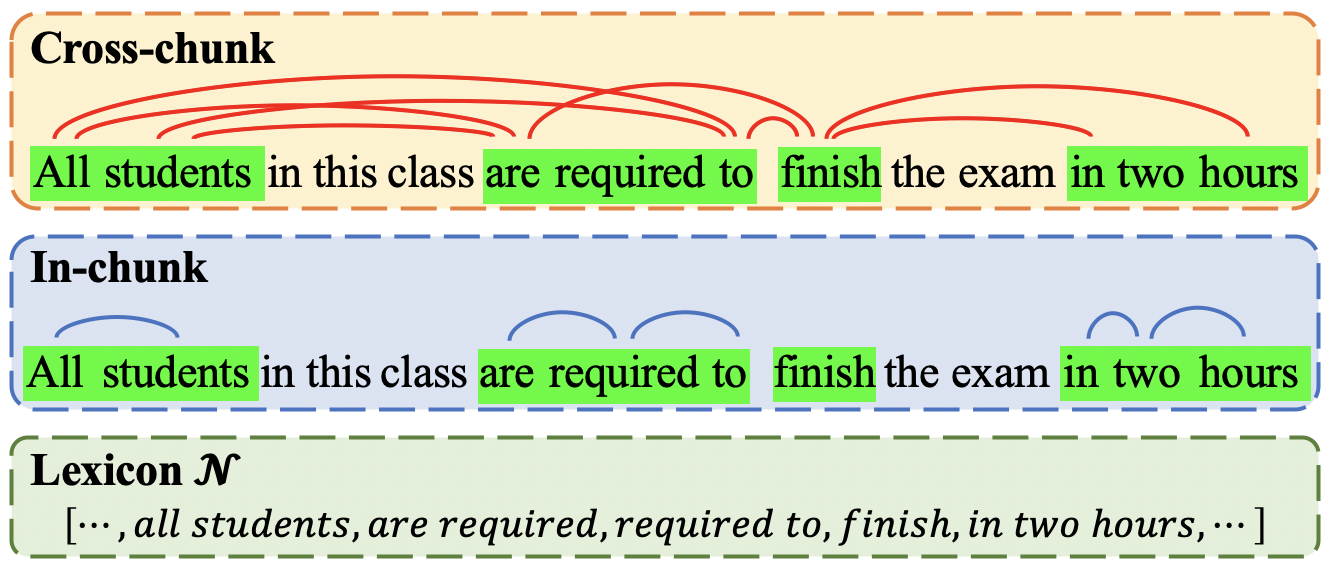}
    \caption{Examples of the two types of edges for building the graph in an input sentence,
    in which chunks (n-grams) extracted from the lexicon $\mathcal{N}$
    are highlighted in green;
    example in-chunk and cross-chunk edges are marked in blue and red color, respectively.
    }
    \label{fig:graph}
    \vskip -1em
\end{figure}
 
Then for graph building, given an input sentence $\mathcal{X}$, we find all the n-grams in $\mathcal{X}$ that 
appear in $\mathcal{N}$. A {\it chunk} is either a n-gram that does not overlap with other n-grams or a text span that covers multiple overlapping n-grams.
For example, in Figure \ref{fig:graph}, we find four chunks (i.e., ``\textit{all students}'', ``\textit{are required to}'', ``\textit{finish}'', and ``\textit{in two hours}'') in the example sentence according to the lexicon $\mathcal{N}$ (the chunks are highlighted in green).
In these chunks, ``\textit{all students}'', ``\textit{finish}'', and ``\textit{in two hours}'' are non-overlapping n-grams included in the lexicon and ``\textit{are required to}'' is a text span that covers the overlapping n-grams ``\textit{are required}'' and ``\textit{required to}''.
%
In most cases, the adjacent words within the same chunk tend to have a strong word-word relation in terms of co-occurrence, and thus we can build the graph and its adjacency matrix accordingly.
%

Based on the chunks, we construct the graph by two types of edges, i.e., the in-chunk and cross-chunk ones:
the first type is to model local word pairs, and the graph includes edges
between any two adjacent words within the same chunk.
For example, as shown in Figure \ref{fig:graph}, the in-chunk edges (blue lines) for the chunk ``\textit{in two hours}'' are ``(\textit{in, two})''
and ``(\textit{two hours})''.
The second type is to model cross chunk word pairs, which are built from any two adjacent chunks with
%
%
the starting and ending words in the two chunks connected.
%
The motivation of using the starting and ending words is that English phrases tend to be head-initial (e.g., verb phrase such as ``\textit{buy some books}'') or head-final (e.g., adjective phrase such as ``\textit{red apples}'') in many cases. E.g.,
for the two chunks ``\textit{all students}'' and ``\textit{are required to}'' in Figure \ref{fig:graph}, the corresponding cross-chunk edges (red lines) are ``\textit{(all, are)}'', ``\textit{(all, to)}'', ``\textit{(students, are)}'', and ``\textit{(students, to)}''.
%
%
The graph is equivalent to the adjacency matrix $\mathcal{A}$, where $a_{i,j}=1$ if there is an edge between $x_i$ and $x_j$ in the graph or $i=j$, and $a_{i,j}=0$ otherwise.\footnote{We do not distinguish the two types of edges in $\mathcal{A}$.}

\subsection{The Attentive GCN}

When learning from a graph, conventional GCN models treat all word pairs from the graph equally, and thus are unable to account for the possibility that the contribution of different 
$x_j$ on $x_i$ may vary. 
Particularly for our graph built from chunks, it is important to be able to distinguish different word pairs because all the chunks and the graph are constructed automatically without a dependency parser.
Therefore, we apply an attention mechanism to the adjacency matrix and adapt Eq. (\ref{eq: gcn}) used in the normal GCN for our A-GCN by replacing the $a_{i,j} \in \{0, 1\}$ by a weight $p^{(l)}_{i,j} \in (0,1)$.
For each $x_i$ and all its associated $x_j$, the weight $p^{(l)}_{i,j}$ for this word pair is computed by
\begin{equation}
\setlength\abovedisplayskip{6pt}
\setlength\belowdisplayskip{6pt}
    p^{(l)}_{i,j} = \frac{a_{i,j} \cdot exp(\mathbf{h}^{(l-1)}_{i} \cdot \mathbf{W}^{(l)}_{pos} \cdot \mathbf{h}^{(l-1)}_{j})}
                    {\sum^{n}_{j = 1} a_{i,j} \cdot exp(\mathbf{h}^{(l-1)}_{i} \cdot \mathbf{W}^{(l)}_{pos} \cdot \mathbf{h}^{(l-1)}_{j})}
\end{equation}
where $\mathbf{W}^{(l)}_{pos}$ models the positional relation (i.e., \textit{left}, \textit{right}, or \textit{self}) between $x_i$ and $x_j$ and it has three choices, i.e., $\mathbf{W}^{(l)}_{left}$, $\mathbf{W}^{(l)}_{right}$, and $\mathbf{W}^{(l)}_{self}$ for different $i$ and $j$,\footnote{For example, $\mathbf{W}^{(l)}_{pos} = \mathbf{W}^{(l)}_{left}$, if $j < i$.} with each of them a trainable square matrix in the $l$-th layer of A-GCN.
%
%

\subsection{Supertagging with A-GCN}

To conduct supertagging with A-GCN, we firstly obtain the hidden vector $\mathbf{h}^{(0)}_{i}$ for $x_{i}$ from BERT \cite{devlin-etal-2019-bert} to feed into the first GCN layer.
Upon receiving the encoding results from A-GCN, the following supertagging process is straightforward:
each $\mathbf{h}^{(L)}_{i}$ is obtained from the last A-GCN layer and aligned with the output by $\mathbf{o}_{i} = \mathbf{W}_{d} \cdot \mathbf{h}^{(L)}_{i}$, where $\mathbf{W}_{d}$ is
a trainable matrix for the alignment.
Then, a \textit{softmax} decoder is used to predict the supertag $\hat{y}_{i}$ for $x_{i}$:
\begin{align} \label{eq:softmax}
\setlength\abovedisplayskip{6pt}
\setlength\belowdisplayskip{6pt}
    \hat{y}_{i} = \arg \max \frac{exp(\mathbf{o}^{t}_i)}
    {\sum_{t=1}^{|\mathcal{T}|} exp(\mathbf{o}^{t}_{i})} 
\end{align}
where $\mathcal{T}$ denotes the set with all CCG categories and $o^{t}_{i}$ the value at dimension $t$ in $\mathbf{o}_{i}$.
%
%
%

\section{Experiments}

\subsection{Settings}

\begin{table}[t]
\begin{center}
\begin{tabular}{l | r r r }
    \toprule
    & \textbf{Train} & \textbf{Dev} & \textbf{Test} \\
    \midrule
    Section No.
    & 2-21 & 0 & 23 \\
    \midrule
    Sentence \# 
    & 39,604 & 1,913 & 2,407
    \\
    Word \#
    & 929,552 & 45,422 & 55,371 
    \\
    \bottomrule
\end{tabular}
\vspace{-0.3cm}
\end{center}
\caption{The train/dev/test splits of English CCGBank and the statistics of sentences and words in them.
}
\label{tab: dataset details}
\end{table}

\begin{table}[t]
\begin{center}
\begin{tabular}{l | c}
    \toprule
    \multirow{1}{*}{\textbf{Hyper-parameters}}
    & \textbf{Values}  \\
    \midrule
    Batch Size & 16, 32 \\
    Drop-out Rate & 0.2 \\
    Learning Rate & 3e-5, 2e-5, 1e-5, 5e-6 \\
    Max Sentence Length & 300 \\
    Random Seed & 42 \\
    Training Epoch & 50 \\
    Warm-up Rate & 0.1, 0.2 \\
    \bottomrule
\end{tabular}
\end{center}
\vspace{-0.3cm}
\caption{
The list of hyper-parameters tested in our experiments.
We run all models with the combination of those hyper-parameters and use the one achieving the highest supertagging results in our final experiments.
}
\label{tab: hyper-parameter}
\vskip -1em
\end{table}

We run experiments on the English CCGbank \cite{hockenmaier-steedman-2007-ccgbank}\footnote{The official dataset is obtained from \url{https://catalog.ldc.upenn.edu/LDC2005T13}.} and
follow \citet{clark-curran-2007-wide} to split it into train/dev/test sets,
whose statistics (sentence and word numbers) are reported in Table \ref{tab: dataset details}.
To construct n-gram lexicon $\mathcal{N}$ for building the edges in our graph, we perform 
PMI
on the training set of CCGbank to extract n-grams whose length is between $[1,5]$, with the threshold of the PMI score set to $0$.
%
%
%
%
%
%
For the encoder, we try both cased and uncased BERT-Large \cite{devlin-etal-2019-bert} with their default settings (e.g., 
24 layers of self-attentions in 1024 dimensional hidden vectors)\footnote{We download the pre-trained BERT models from \url{https://github.com/google-research/bert}.} and used two layers for A-GCN.
%
%
To obtain CCG parse from the generated supertags, we adopt the parsing algorithm used in EasyCCG \cite{lewis-steedman-2014-ccg}.
%
%
%
We follow previous studies \cite{lewis-steedman-2014-ccg, lewis-etal-2016-lstm, yoshikawa-etal-2017-ccg} to use the most frequent 425 supertags as the tag set and evaluate our model on both the tagging accuracy and the labeled F-scores (LF) of the dependencies converted from CCG parse\footnote{We use the ``\textit{generate}'' script from C\&C parser \cite{clark-curran-2007-wide} to convert CCG parse results to their corresponding dependencies.}.

For other hyper-parameter settings, we test their values as shown in
Table \ref{tab: hyper-parameter} when training our models.
We tried all combinations of them for each model and use the one achieving the highest supertagging results in our final experiments.
Note that, with the best hyper-parameters, the best performance is achieved with warm-up rate 0.1, batch size 16, and learning rate 1e-5.
%

\subsection{Results}

\begin{table}[t]
\begin{center}
\begin{small}
\begin{tabular}{l | C{1.0cm} | C{1.0cm} C{1.0cm}}
    \toprule
    \multirow{1}{*}{\textbf{Models}}
    & \textbf{PARM} & \textbf{TAG} & \textbf{LF} \\
    \midrule
    BERT-Cased 
    & 335M
    & 96.04 & 90.31 \\
    \ \ \  + A-GCN (Full)
    & 343M
    & 95.93 & 90.13 \\
    \ \ \   + A-GCN (Chunk)
    & 343M
    & \textbf{96.11} & \textbf{90.49} \\
    \midrule
    BERT-Uncased 
    & 337M
    & 96.06 & 90.37 \\
    \ \ \  + A-GCN (Full)
    & 345M
    & 95.99 & 90.21 \\
    \ \ \   + A-GCN (Chunk)
    & 345M
    & \textbf{96.17} & \textbf{90.60} \\
    \bottomrule
\end{tabular}
\end{small}
\end{center}
\vspace{-0.3cm}
\caption{
Results (supertagging accuracy and labeled $F$-scores)
of different models with BERT-Large encoder on the development set of CCGbank. 
``PARM'' is the number of trainable parameters in the models;
``Full'' uses the fully connected graph and ``Chunk'' uses the graph built based on chunks.}
\label{tab: results}
\vskip -1em
\end{table}

To explore the effectiveness of our approach, we run CCG taggers with and without A-GCN,
and try two ways to construct the graph: one is a fully connected GCN where edges are built between every two words; the other is our proposed approach with the
chunk-based graph.
%
%
Experimental results on supertagging accuracy (\textsc{Tag}) and labeled F-scores (\textsc{LF}) for parsing on the development set of CCGbank
are reported in Table \ref{tab: results},
with the number of trainable parameters of all models also presented.

The experiments show that, for both cased and uncased BERT encoders, the proposed chunk A-GCN works the best in terms of both supertagging accuracy and parsing results.
%
%
%
In contrast, Full A-GCN has inferior performance to the BERT baselines.
%
This contrast shows the importance of appropriate construction of the graphs fed into A-GCN,
since the fully connected graph with all words associated with one another may introduce noise 
word relations and thus yield bad performance.
%
%
%

Furthermore, we run our models with uncased BERT encoder on the test set and compare the 
performance
with previous studies on both supertagging and parsing.
Table \ref{tab: sota} shows the results, where the studies marked by $\dagger$ use the same parser (i.e., the EasyCCG parser) to generate CCG trees from supertags.
%
%
Among the previous studies, \citet{stanojevic-steedman-2019-ccg} performed CCG parsing directly without the suppertagging step, whereas the rest all did supertagging first.
Regardless of this difference, our approach performs the best on CCGbank in both supertagging accuracy and parsing LF.
%

\begin{table}[t]
\begin{center}
\begin{small}
\begin{tabular}
    {L{4.5cm} | C{0.8cm} C{0.8cm}}
    \toprule
    \textbf{Models} & \textbf{TAG} & \textbf{LF} \\
    \midrule
    \newcite{lewis-steedman-2014-improved} 
    & 91.3~~~ & 86.11 \\
    \newcite{xu-etal-2015-ccg} 
    & 93.00 & 87.07 \\
    \newcite{lewis-etal-2016-lstm} 
    & 94.7~~~ & 88.1~~~ \\
    \newcite{vaswani-etal-2016-supertagging} 
    & 94.24 & 88.32 \\
    \newcite{yoshikawa-etal-2017-ccg}
    & - & 90.4~~~ \\
    \newcite{clark2018semi} 
    & 96.1~~~ & - \\
    \newcite{stanojevic-steedman-2019-ccg}
    & 95.4~~~ & 90.5~~~ \\
    \midrule
    EasyCCG$\dagger$
    & - & 86.14 \\
    \midrule
    BERT$\dagger$
    & 96.06 & 90.34 \\
    BERT + A-GCN (Full)$\dagger$
    & 95.91 & 90.20 \\
    BERT + A-GCN (Chunk)$\dagger$
    & \textbf{96.25} & \textbf{90.58} \\
    \bottomrule
\end{tabular}
\end{small}
\end{center}
\vspace{-0.3cm}
\caption{Comparison of our models with uncased BERT encoder and previous studies on the test set of CCGbank.
Models with ``$\dagger$'' use the EasyCCG parser to generate CCG parse trees from the predicted supertags.}
\label{tab: sota}
\vskip -1em
\end{table}

\subsection{Ablation Study}

%
We conduct an ablation study to explore the effect of the two types of edges and the attention mechanism on our best model.
%
%
The supertagging and parsing results of models with different configurations are reported in Table \ref{tab: ablation},
where the results are categorized into four groups.
The first group (ID 1) is the results of the best performing model where all settings are activated;
the second (ID 2-3) is the ablation of either in-chunk or cross-chunk edges with attention;
the third (ID 4-6) is the result of using normal GCN without the attention mechanism;
and the last group (ID 7) is the baseline model where none of the three settings is activated.
%
%

The results show that the model performance drops when either part is ablated (ID 1 vs. ID 2-6).
Specifically,
removing attention 
significantly hurts the performance, where all configurations without attention (ID 4-6) shows worse-than-baseline (ID 7) results; this observation confirms the importance of applying attention on GCN.
One possible explanation to this phenomenon could be that considerable noises are introduced to the graph because the edges in our graph are derived from chunks and they do not follow syntax in most cases;
thus, it is crucial to assign weights to the edges and not treat them with equally.
%
Interestingly, comparing the two types of edges, models with cross-chunk edges yield much higher results
than the ones with in-chunk edges only when the attention is not used (ID 5 vs. ID 6),
while it is slightly better when attention is applied (ID 2 vs. ID 3).
This comparison suggests that in-chunk edges could introduce more noise than cross-chunk edges.
So that
when the attention is not used (ID 6), the model fails to weight the edges and results in a significant drop on its performance;
On the contrary, when the attention is applied (ID 3),
our model is able to even the performance of models with in-chunk and cross-chunk edges, which confirms that weighting is essential in selecting useful information for CCG supertagging.

\begin{table}[t]
\begin{center}
\begin{small}
\begin{tabular}
    {C{0.3cm} | C{1.15cm} C{1.6cm} C{1cm} | C{0.55cm} C{0.55cm}}
    \toprule
    \multirow{2}{*}{\textbf{ID}} & \multicolumn{3}{c|}{\textbf{Settings}} & \multirow{2}{*}{\textbf{Tag}} & \multirow{2}{*}{\textbf{LF}} \\
    & In-chunk & Cross-chunk & Attention &  &  \\
    \midrule
    1 & $\surd$ & $\surd$ & $\surd$
    & \textbf{96.25} & \textbf{90.58} \\
    \midrule
    2 & $\texttimes$ & $\surd$ & $\surd$
    & 96.18 & 90.49 \\
    3 & $\surd$ & $\texttimes$ & $\surd$
    & 96.11 & 90.41 \\
    \midrule
    4 & $\surd$ & $\surd$ & $\texttimes$
    & 87.26 & 81.96 \\
    5 & $\texttimes$ & $\surd$ & $\texttimes$
    & 94.92 & 89.75 \\
    6 & $\surd$ & $\texttimes$ & $\texttimes$
    & 89.67 & 84.39 \\
    \midrule
    7 & $\texttimes$ & $\texttimes$ & $\texttimes$
    & 96.06 & 90.34 \\
    \bottomrule
\end{tabular}
\end{small}
\end{center}
\vspace{-0.3cm}
\caption{Experimental results of models with uncased BERT-Large encoder on the test set of CCGbank, where the in-chunk, cross-chunk edges or the attention mechanism in our A-GCN module is ablated. 
}
\label{tab: ablation}
\vskip -1em
\end{table}

\section{Conclusion}

%

In this paper, we propose A-GCN for CCG supertagging, with its graph built from chunks extracted from a lexicon.
We use two types of edges for the graph, namely, in-chunk and cross-chunk edges for word pairs within and across chunks, respectively,
and propose an attention mechanism 
%
to distinguish the important word pairs according to their contribution to CCG supertagging.
%
Experimental results and the ablation study on the English CCGbank demonstrate the effectiveness of our approach to CCG supertagging, where state-of-the-art performance is obtained on both CCG supertagging and parsing.
Further analysis is performed to investigate using different types of edges,
which reveals their quality and confirms the necessity of introducing attention to GCN for CCG supertagging.

\section*{Acknowledgements}

This work is supported by The Chinese University of Hong Kong (Shenzhen) under University Development Fund UDF01001809.


\bibliography{emnlp2020}
\bibliographystyle{acl_natbib}

\appendix

\vspace{0.2cm}

\section*{Appendix A: Example Sentences with Extracted Chunks}

\begin{figure}[h]
    \centering
    \includegraphics[width=0.48\textwidth, trim=0 10 0 0]{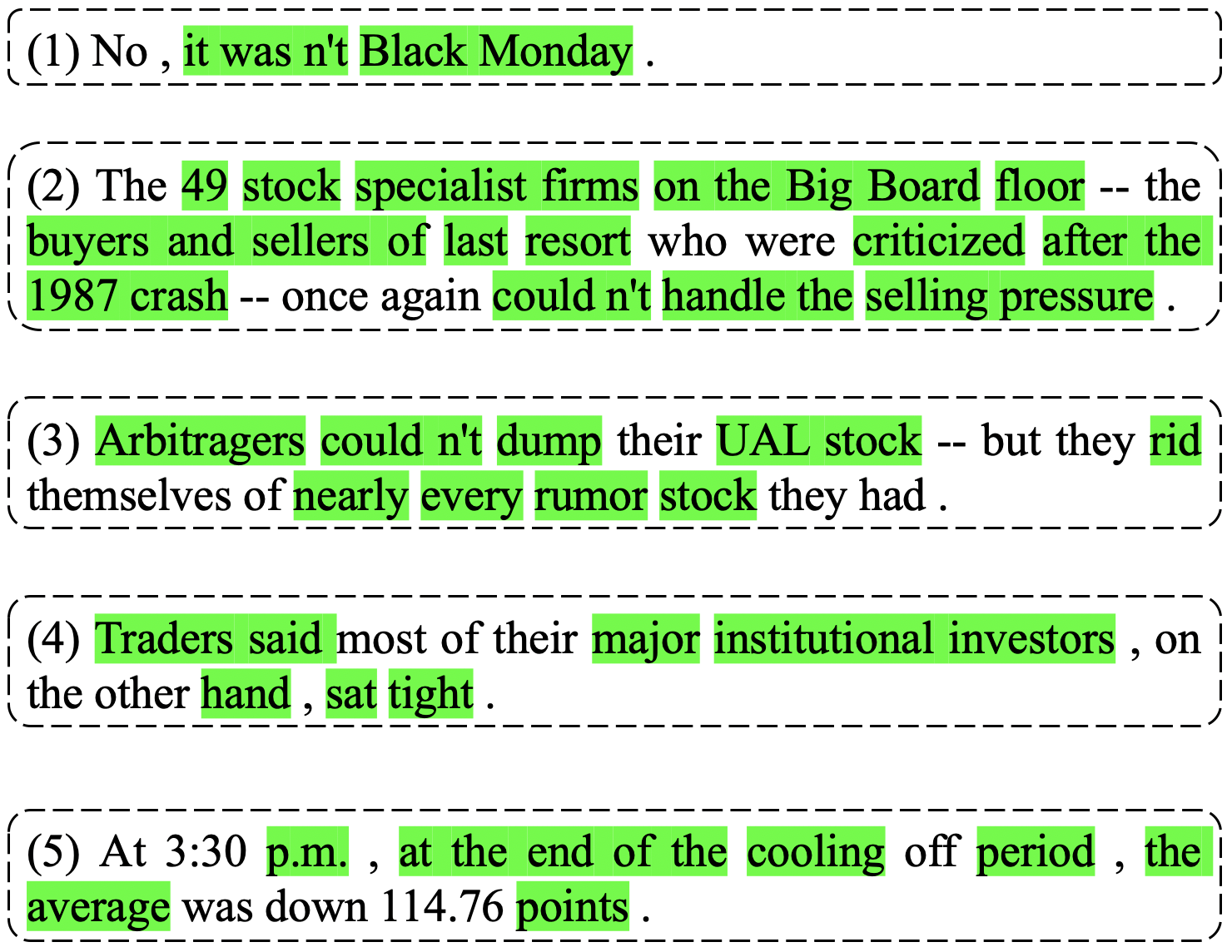}
    \caption{Example sentences with the chunks extracted from the lexicon $\mathcal{N}$ highlighted in green.
    }
    \label{fig:example}
    \vskip -0.5em
\end{figure}

In the main experiments, we use the lexicon obtained from the training set of the English CCGbank to extract chunks in each sentence, where the chunks are used to build the graph.
Figure \ref{fig:example} shows five example sentences in which the extracted chunks are highlighted in green.
We report more examples in the supplemental materials.


\section*{Appendix B: Example Suppertagging and Parsing Results}

Figure \ref{fig:example parse} shows the CCG supertagging and parsing results of EasyCCG\footnote{We use the implementation from CCGweb \cite{evang-etal-2019-ccgweb} at \url{https://ccgweb.phil.hhu.de/}.} and our approach (i.e., BERT + A-GCN (Chunk)) on two example sentences.
In the figure, the correct and incorrect supertags are represented by green and red color, respectively.
We report more CCG parsing results of our approach in the supplemental materials.

\definecolor{darkgreen}{rgb}{0.36, 0.6, 0.25}

\begin{figure*}[h]
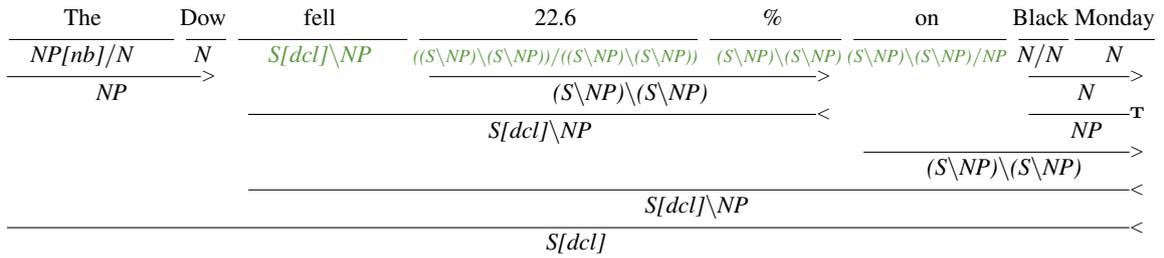

    %
    %
    \begin{subfigure}[h]{1.0\textwidth}
    \raggedright
    \hspace{0cm} \textbf{EasyCCG} \\
    \begin{center}
    \cgex{7}{
    The & Dow & Jones & industrials & closed & at & 2569.26 \\
    \cglines{7}\\
    \cgf{NP[nb]\fs N} & \textcolor{red}{\cgf{(N\fs N)\fs (N\fs N)}} & \cgf{N\fs N} & \cgf{N} & \textcolor{red}{\cgf{S[dcl]\bs NP}} & \textcolor{red}{\cgf{((S\bs NP)\bs (S\bs NP))\fs NP}} & \cgf{N} \\
    & \cgline{2}{\cgfa} & & & & \cgline{1}{\cgtr} \\
    & \cgres{2}{N\fs N} & & & & \cgres{1}{NP} \\ 
    & \cgline{3}{\cgfa} & & \cgline{2}{\cgfa} \\
    & \cgres{3}{N} & & \cgres{2}{(S\bs NP)\bs (S\bs NP)} \\
    \cgline{4}{\cgfa} & \cgline{3}{\cgba} \\
    \cgres{4}{NP} & \cgres{3}{S[dcl]\bs NP} \\
    \cgline{7}{\cgba} \\
    \cgres{7}{S[dcl]} \\
    \\
    \vspace{0.3cm}
    \hspace{-1.cm} \textbf{Our Approach} \\
    The & Dow & Jones & industrials & closed & at & 2569.26 \\
    \cglines{7} \\
    \cgf{NP[nb]\fs N} & \textcolor{darkgreen}{\cgf{N\fs N}} & \cgf{N\fs N} & \cgf{N} & \textcolor{darkgreen}{\cgf{(S[dcl]\bs NP)\fs PP}} & \textcolor{darkgreen}{\cgf{PP\fs NP}} & \cgf{N} \\
    & & \cgline{2}{\cgfa} & & & \cgline{1}{\cgtr} \\
    & & \cgres{2}{N} & & & \cgres{1}{NP} \\ 
    & \cgline{3}{\cgfa} & & \cgline{2}{\cgfa} \\
    & \cgres{3}{N} & & \cgres{2}{PP} \\
    \cgline{4}{\cgfa} & \cgline{3}{\cgba} \\
    \cgres{4}{NP} & \cgres{3}{S[dcl]\bs NP} \\
    \cgline{7}{\cgba} \\
    \cgres{7}{S[dcl]} \\
    }
    \end{center}
    \caption{}
    \end{subfigure}
    %
    %
    \begin{small}
    \begin{subfigure}[h]{0.98\textwidth}
    \hspace{0.1cm} {\normalsize \textbf{EasyCCG}} \\
    \begin{center}
    \cgex{8}{
    The & Dow & fell & 22.6 & \% & on & Black & Monday \\
    \cglines{8}\\
    \cgf{NP[nb]\fs N} & \cgf{N} & \textcolor{red}{{\scriptsize\cgf{((S[dcl]\bs NP)\fs PP)\fs NP}}} & \textcolor{red}{\cgf{N\fs N}} & \textcolor{red}{\cgf{N}} & \textcolor{red}{\cgf{PP\fs NP}} & \cgf{N\fs N} & \cgf{N} \\
    \cgline{2}{\cgfa} & & \cgline{2}{\cgfa} & & \cgline{2}{\cgfa} \\
    \cgres{2}{NP} & & \cgres{2}{N} & & \cgres{2}{N} \\ 
    & & & \cgline{2}{\cgtr} & & \cgline{2}{\cgtr} \\
    & & & \cgres{2}{NP} & & \cgres{2}{NP} \\
    & & \cgline{3}{\cgfa} & \cgline{3}{\cgfa} \\
    & & \cgres{3}{(S[dcl]\bs NP)\fs PP} & \cgres{3}{PP} \\
    & & \cgline{6}{\cgfa} \\
    & & \cgres{6}{S[dcl]\bs NP} \\
    \cgline{8}{\cgba} \\
    \cgres{8}{S[dcl]} \\
    \\
    \vspace{0.3cm}
    {\normalsize \textbf{Our Approach}} \\
    The & Dow & fell & 22.6 & \% & on & Black & Monday \\
    \cglines{8}\\
    \cgf{NP[nb]\fs N} & \cgf{N} & \textcolor{darkgreen}{\cgf{S[dcl]\bs NP}} & \textcolor{darkgreen}{{\scriptsize\cgf{((S\bs NP)\bs (S\bs NP))\fs ((S\bs NP)\bs (S\bs NP))}}} &
    \hspace{0.1cm} \textcolor{darkgreen}{{\scriptsize\cgf{(S\bs NP)\bs (S\bs NP)}}} & \textcolor{darkgreen}{{\scriptsize\cgf{(S\bs NP)\bs (S\bs NP)\fs NP}}} & \cgf{N\fs N} & \cgf{N} \\
    \cgline{2}{\cgfa} & & \cgline{2}{\cgfa} & & \cgline{2}{\cgfa} \\
    \cgres{2}{NP} & & \cgres{2}{(S\bs NP)\bs (S\bs NP)} & & \cgres{2}{N} \\ 
    & & \cgline{3}{\cgba} & & \cgline{2}{\cgtr} \\
    & & \cgres{3}{S[dcl]\bs NP} & & \cgres{2}{NP} \\
    & & & & & \cgline{3}{\cgfa} \\
    & & & & & \cgres{3}{(S\bs NP)\bs (S\bs NP)} \\
    & & \cgline{6}{\cgba} \\
    & & \cgres{6}{S[dcl]\bs NP} \\
    \cgline{8}{\cgba} \\
    \cgres{8}{S[dcl]} \\
    }
    \end{center}
    \caption{}
    \end{subfigure}
    \end{small}
    \caption{
    The CCG supertagging and parsing results of EasyCCG and our approach (i.e., BERT + A-GCN (Chunk)) on the examples, where the correct and incorrect supertags are represented in green and red color, respectively.
    }
    \label{fig:example parse}
\end{figure*}

\end{document}